\documentclass[aps,twocolumn,preprintnumbers,groupedaddress,nofootinbib]{revtex4-1}
\pdfoutput=1
\usepackage[usenames,dvipsnames,table]{xcolor}
\usepackage{graphicx} 
\usepackage{amsmath,amssymb,amsthm,multirow,array,bm,bbm,esint}
\usepackage[mathscr]{eucal}
\usepackage[bbgreekl]{mathbbol}
\usepackage{epsf,amsfonts}
\usepackage{physics}
\usepackage{dsfont,mathtools}
\usepackage{slashed}
\usepackage{hyperref}
\usepackage{mathdots}
\usepackage{subcaption}
\usepackage{booktabs}
\usepackage{microtype}
\usepackage[section]{placeins} 
\usepackage{float}
\usepackage{algorithm}
\usepackage{algpseudocode}
\usepackage{algpseudocode}
\usepackage{tikz}
\usepackage[normalem]{ulem}

\newcommand{\SC}{\mathrm{SC}}
\newcommand{\AS}{\mathrm{AS}}
\newcommand{\VA}{\mathrm{VA}}
\newcommand{\CNR}{\mathrm{CNR}}

\theoremstyle{definition}

\theoremstyle{plain}

\renewcommand{\thesubsection}{\thesection.\arabic{subsection}}
\usepackage{titlesec}
\titleformat{\subsection}{\normalfont\large\bfseries}{\thesubsection}{1em}{}

\usepackage{titletoc}
\titlecontents{section}[0em]{\vspace{0.5em}\normalfont\bfseries}
    {\thecontentslabel\quad}{}{\titlerule*[0.5pc]{.}\contentspage}
\titlecontents{subsection}[1.5em]{\normalfont}
    {\thecontentslabel\quad}{}{\titlerule*[0.5pc]{.}\contentspage}

\begin{document}

\title{Quantifying constraint hierarchies in Bayesian PINNs via per-constraint Hessian decomposition}

\author{Filip Landgren}
\affiliation{School of Mathematical Sciences and STAG Research Centre, University of Southampton, Highfield, Southampton SO17 1BJ, UK}
\email{f.landgren@soton.ac.uk}

\begin{abstract}
Bayesian physics-informed neural networks (B-PINNs) merge data with governing equations to solve differential equations under uncertainty.  
However, interpreting uncertainty and overconfidence in B-PINNs requires care due to the poorly understood effects the physical constraints have on the network; overconfidence could reflect warranted precision, enforced by the constraints, rather than miscalibration.
Motivated by the need to further clarify how (individual) constraints shape these networks, we introduce a scalable, matrix-free Laplace framework that decomposes the posterior Hessian into contributions from each constraint and provides metrics to quantify their relative influence on the loss landscape. Applied to the Van der Pol equation, our method tracks how constraints sculpt the network’s geometry and shows, directly through the Hessian, how changing a single loss weight non-trivially redistributes curvature and effective dominance across the others.
\end{abstract}

\maketitle

\section{Introduction}\label{sec:intro}
Bayesian physics-informed neural networks (B-PINNs) integrate observational data with physical laws under a probabilistic framework, enabling uncertainty quantification (UQ) in solving differential equations (see \cite{Psaros_2023} for a survey on UQ in scientific machine learning, and \cite{graf2022errorawarebpinnsimprovinguncertainty} for a recent study of UQ for B-PINNs). However, UQ in B-PINNs exhibits subtle pathologies: dominant constraints can induce apparent overconfidence by restricting the solution manifold, while the emergent influence of each constraint transcends its nominal loss weight due to curvature interactions. As emphasized in [first pinn paper], physical constraints in B-PINNs can collapse degrees of freedom in the solution space, leading to, by design,  regions of low predictive variance that manifest as apparent overconfidence
This phenomenon arises naturally from enforcing the governing physics and differs fundamentally from typical sources of overconfidence, such as overfitting or miscalibration in conventional neural networks. A key challenge is to quantitatively understand how physical constraints carve out the loss landscape, shaping the posterior curvature in specific directions. In the present note, we build upon this by developing a per-component Hessian-based framework that quantifies how individual constraints shape the loss landscape and elucidates their influence on Bayesian uncertainty estimates.

Loss weights offer only crude control over the enforcement of the corresponding loss term; if we want to more strictly enforce a physical constraint, we cannot necessarily expect it to be achieved by just increasing the corresponding weights, as that might break something elsewhere in the network with propagating effects. 
In this work we explicitly demonstrate the non-trivial implicit relation between the physical constraints.  Furthermore, we demonstrate, through the lens of the Hessian, how adjusting loss weights significantly alters the hierarchical influence of constraints on the network.

The algorithmic pipeline builds on well-known ingredients: Laplace approximation for Bayesian neural nets, Hessian-vector products (HVPs) \cite{Pearlmutter1994FastHessian}, Lanczos for top eigenpairs, conjugate-gradient (CG) solves, and optional Gauss-Newton surrogates, applied in a new context and combination. Following standard Laplace practice \cite{daxberger2022laplacereduxeffortless, Ritter2018LaplaceScalable}, we freeze the B-PINN at its variational mean/MAP to obtain a deterministic network and evaluate the posterior Hessian at that point; this step, is essential for stable matrix-free curvature estimation.

Our contribution leverages this determinized Hessian to decompose posterior curvature per constraint (data, PDE residual, initial condition (IC), and boundary conditions (BC)) and to define four new metrics: Spectral Contribution (SC), Alignment Score (AS), Variance Attribution (VA), and Condition-Number Ratio (CNR), that probe how each physical condition carves the loss landscape and induces physics-driven overconfidence. In other words, while the numerical primitives are established (e.g., as in toolkits like PyHessian \cite{Yao2019PyHessian}), the Bayesian, per-constraint attribution of posterior precision and the resulting hierarchy analysis are new, and the freezing step is central because it makes these measurements coherent and reproducible.
We demonstrate this framework on the Van der Pol oscillator, revealing how hierarchies shift with stiffness and weighting-insights not evident from weights alone. This methodology addresses B-PINN subtleties, providing tools to diagnose the effects physical constraints have on the surrogate model.

In B-PINNs, the negative log-posterior is:
\begin{multline}
\mathcal{U}(\theta) = \lambda_\text{data} \mathcal{L}_\text{data} + \lambda_\text{PDE} \mathcal{L}_\text{PDE} \\+ \lambda_\text{IC} \mathcal{L}_\text{IC} + \lambda_\text{BC} \mathcal{L}_\text{BC} + \mathcal{L}_\text{prior},
\end{multline}
where $\mathcal{L}$ are the loss weights of the corresponding source of loss $\mathcal{L}$ that the network tries to minimize \cite{Yang2021BPINNs}.

We employ a Laplace approximation $\mathcal{N}(\hat{\theta}, H^{-1})$ around the variational mean $\hat{\theta}$, with $H = \nabla^2 \mathcal{U}(\hat{\theta})$ \cite{daxberger2022laplacereduxeffortless,Ritter2018LaplaceScalable}. This captures local curvature, linking to loss landscapes \cite{li2018visualizinglosslandscapeneural}.

Taken together, the results offers a ranking that reveals when loss weights diverge from information dominance, explicitly demonstrating from the point of the view of the Hessian, as far as we are aware, for the first time, the non-trivial reallocation of constraint influence under weight adjustments, where tweaking one term alters the effective dominance of others through shared parameter subspaces and curvature couplings.

\bigskip

\textbf{Relation to prior work.}
Deterministic analyses of PINN loss landscapes document ill-conditioning and, in some cases, per-component spectra to motivate optimizers \cite{Rathore2024PINNLandscapes, Krishnapriyan2021FailurePINNs}, and complementary work visualizes high-dimensional landscapes \cite{geniesse2024visualizinglossfunctionstopological,li2018visualizinglosslandscapeneural}.  
In parallel, scalable Laplace methods for generic neural networks deliver local posterior approximations but treat the objective monolithically \cite{Ritter2018LaplaceScalable, daxberger2022laplacereduxeffortless,Daxberger2021LaplaceRedux}.  
UQ surveys and B-PINN formulations advance probabilistic treatments of physics-informed models \cite{Psaros_2023,Yang2021BPINNs,Raissi2019PINN,Jagtap2020XPINNs} but do not decompose Hessians by physics term.  
Work on gradient stiffness and weighting pathologies in PINNs \cite{wang2020understandingmitigatinggradientpathologies,Ji2021StiffPINN} motivates going beyond nominal weights.  
Our contribution bridges these threads by introducing a Laplace-based, per-cinstraint Hessian decomposition and associated metrics (SC/AS/VA/CNR) that quantify constraint dominance and cross-constraint coupling in B-PINNs.

\section{Methodology and experimental setup}
 The approach builds on the Laplace approximation to linearize the posterior and decompose the Hessian into per-constraint contributions, enabling the computation of metrics that reveal "impact" the individual constraints has on the loss landscape. While the metrics are empirically motivated rather than formally derived, they are designed to capture distinct aspects of how constraints shape the posterior geometry, drawing inspiration from related concepts in neural network optimization and Hessian analysis.
A key subtlety in B-PINN UQ is that physical constraints can tighten feasible parameter directions, elevating eigenvalues of the Hessian $H$ along those modes and thereby reducing predictive variance via terms like $J_x^\top H^{-1} J_x$, where $J_x$ is the Jacobian of the network output at point $x$. This manifests as apparent overconfidence, which is desirable if the physics and noise models are accurate but can otherwise obscure misspecification \cite{daxberger2022laplacereduxeffortless}. Our framework isolates which constraints drive this precision boost.
Exploiting the linearity of the Hessian, we decompose
\begin{equation} H_{\text{tot}}=\sum_{c \in \{\text{data, pde,ic,bc} \} } \lambda_c H_c + H_{\text{prior}},
\end{equation}
where $H_c=\nabla^2 \mathcal{L}_c(\hat{\theta})$ at the variational mean $\hat{\theta}$. We never materialize dense matrices; instead, each $H_c$ is queried via matrix-free Hessian-vector products (HVPs) \cite{Pearlmutter1994FastHessian}, with leading eigenpairs estimated using Lanczos and linear solves via CG with Tikhonov regularization $(H_c + \varepsilon I)$ for stability. In regimes where the full Hessian may be indefinite, we optionally employ a positive semi-definite Gauss-Newton surrogate, as commonly used in second-order optimization \cite{Schraudolph2002GN}.
Let ${(\lambda_j,q_j)}{j=1}^k$ denote the top-$k$ eigenpairs (by absolute magnitude) of $H_{\text{tot}}$. For each constraint $c$, we introduce four diagnostics that quantify how individual constraints (data, PDE, IC, BC) shape the local geometry of a trained B-PINN under a Laplace view. They are heuristic but operational: each captures a distinct way a constraint can dominate curvature and, by extension, local uncertainty.
\begin{itemize}
\item \textbf{Spectral Contribution (SC):} 
\begin{equation}
\SC_c = \frac{\sum_{j=1}^{k} \lambda^{(c)}_j}{\sum_{j=1}^{k} \lambda_j},
\end{equation}
where $\lambda^{(c)}_j$ are the top eigenvalues of $H_c$,  and $\{\lambda_j\}_{j=1}^k$ those of $H_{\mathrm{tot}}$. SC measures how much of the principal curvature budget comes from constraint $c$. High SC signals that $c$ is  responsible for collapsing degrees of freedom in the stiff directions, which is expected to coincide with reduced local predictive variance.

\item \textbf{Alignment Score (AS):} With $g_c=\nabla \mathcal{L}c(\hat{\theta})$ and weights $w_j=\lambda_j/\sum_{i=1}^k\lambda_i$,
\begin{equation}
\AS_c = \sum_{j=1}^{k} w_j \frac{|\langle g_c, q_j \rangle|}{\|g_c\|\,\|q_j\|}.
\end{equation}
This quantifies how well the gradient of $c$ aligns with the total Hessian's eigenvectors, indicating directional influence on optimization and uncertainty; similar alignment metrics appear in gradient pathology studies for PINNs. Weighting by $w_j$ prioritizes directions that most affect optimization and posterior geometry. High AS means $c$ effectively “steers” the principal curvatures; this explains why a term with a modest weight can still dominate uncertainty if it aligns with high-curvature modes. Practically, AS reveals cross-constraint entanglement via overlapping gradient subspaces.
\item \textbf{Variance Attribution (VA):} For a grid $\mathcal{X}$,
\begin{equation}
\VA_c = \frac{1}{|\mathcal{X}|}\sum_{x\in\mathcal{X}} J_x^\top (H_c + \varepsilon I)^{-1} J_x,
\end{equation}
Under the Laplace approximation, predictive variance uses $H_{\mathrm{tot}}^{-1}$. VA is a, diagnostic attribution: we ask how large the variance would be if we retained only curvature from $c$. Low VA indicates that $c$ confers high precision (large curvature) along directions that matter for $f(x)$ on $\mathcal{X}$. Empirically, this connects to our earlier “overconfidence can be warranted” observation: constraints that meaningfully tighten the posterior manifold yield visibly lower VA.
\item \textbf{Condition-Number Ratio (CNR):} 
\begin{equation}
\CNR_c = \kappa(H_c) / \kappa(H_{\text{tot}}),
\end{equation} where $\kappa(\cdot)$ is the condition number (ratio of largest to smallest eigenvalue magnitude). This compares the stiffness introduced by $c$ relative to the full problem.
\end{itemize}

These metrics, while not previously combined in this exact
form, draw from established tools in the literature. 
SC and CNR build on analyses of deep-network Hessian spectra and and ill-conditioning \cite{ghorbani2019investigationneuralnetoptimization, Pennington2017Geometry, sagun2017eigenvalueshessiandeeplearning, Rathore2024PINNLandscapes}.  
AS is motivated by evidence that SGD gradients concentrate in the top Hessian eigenspace \cite{ghorbani2019investigationneuralnetoptimization} and by cosine-based gradient-conflict diagnostics \cite{wang2025gradientalignmentphysicsinformedneural}, but here we weight by eigenvalue magnitude and evaluate each constraint separately.  
VA follows the Laplace predictive-variance formula $J_x^\top H_{\text{tot}}^{-1}J_x$ and extends it by substituting $H_c^{-1}$ to attribute variance to a single constraint, conceptually akin to inverse-Hessian influence functions \cite{koh2020understandingblackboxpredictionsinfluence}. 

Their aggregation into a simple equal-weighted rank provides a visual hierarchy, though the individual values offer the primary insights.
Computationally, each HVP incurs roughly one forward-backward pass cost. Lanczos requires $O(k)$ HVPs for top eigenpairs, while VA involves CG solves per Jacobian row on a modest grid ($\sim 500$ points here), rendering the pipeline scalable for mid-sized B-PINNs (e.g., $\sim 10^4$ parameters). The process is summarized in Algorithm~\ref{alg:hierarchy}.
For the experimental setup, we apply this framework to the Van der Pol oscillator, a nonlinear ODE emblematic of stiff dynamics:
\begin{equation}
\frac{d^2u}{dt^2} - \mu (1 - u^2) \frac{du}{dt} + u = 0, \quad t \in [0,7],
\end{equation}
with initial conditions $u(0)=2$, $du/dt(0)=0$. The solution is constrained by the PDE residual across collocation points and sparse data (20 points, including $ t = 7 $ with $ u(7) \approx 1.6978 $ interpolated from the true $\mu=1$ solution), which serves as a soft endpoint anchor. This setup allows us to probe implicit constraint influences, especially in the \emph{no-BC} case where $\lambda_{\text{BC}}=0$. We generate sparse data from the $\mu=1$ solution but train under varied regimes to probe hierarchy shifts
The B-PINN architecture comprises 3 hidden layers with 50 neurons each, using Tanh activations and Bayesian linear layers implemented via the Blitz library \cite{esposito2020blitzbdl}, which employs Bayes by Backprop for variational inference with a scale-mixture prior ($\sigma_1=0.1$, $\sigma_2=0.1$, $\pi=0.5$).
Training uses Adam optimization (lr=$10^{-3}$) over 2000 epochs, with 5 Monte Carlo samples and KL divergence scaled by $10^{-3}$. Configurations include: \emph{base} ($\mu=1$, all $\lambda=1$), \emph{high-$\mu$} ($\mu=10$), \emph{high-$\lambda_{\text{PDE}}$} ($\lambda_{\text{PDE}}=10$), \emph{low-$\lambda_{\text{PDE}}$} ($\lambda_{\text{PDE}}=0.1$), and \emph{no-BC} ($\lambda_{\text{BC}}=0$). Post-training, we freeze to the variational mean for deterministic Hessian analysis, computing metrics and eigenspectra as described.

\begin{algorithm}[H]
\caption{Constraint-hierarchy analysis for a trained B-PINN}
\label{alg:hierarchy}
\begin{algorithmic}[1]
\Require Trained parameters $\hat{\theta}$; losses $\mathcal{L}_c$; grid $\mathcal{X}$; top-$k$.
\State Freeze Bayesian layers to variational means (analysis-time only).
\For{total and each $c\in\text{data,pde,ic,bc}$}
\State Estimate top-$k$ eigenpairs of $H_c$ (or $H_{\text{tot}}$) via Lanczos using HVPs.
\State Compute $g_c=\nabla \mathcal{L}c(\hat{\theta})$ and Jacobians ${J_x}{x\in\mathcal{X}}$.
\State Compute $\SC_c$, $\AS_c$, $\CNR_c$.
\State Compute $\VA_c$ via CG solves of $(H_c+\varepsilon I)z=J_x$ and average $J_x^\top z$.
\EndFor
\State Aggregate metrics to a rank for visualization (equal weights by default).
\end{algorithmic}
\end{algorithm}

\section{Results}\label{sec:results}

\begin{figure}[t]
\centering
\includegraphics[width=.85\linewidth]{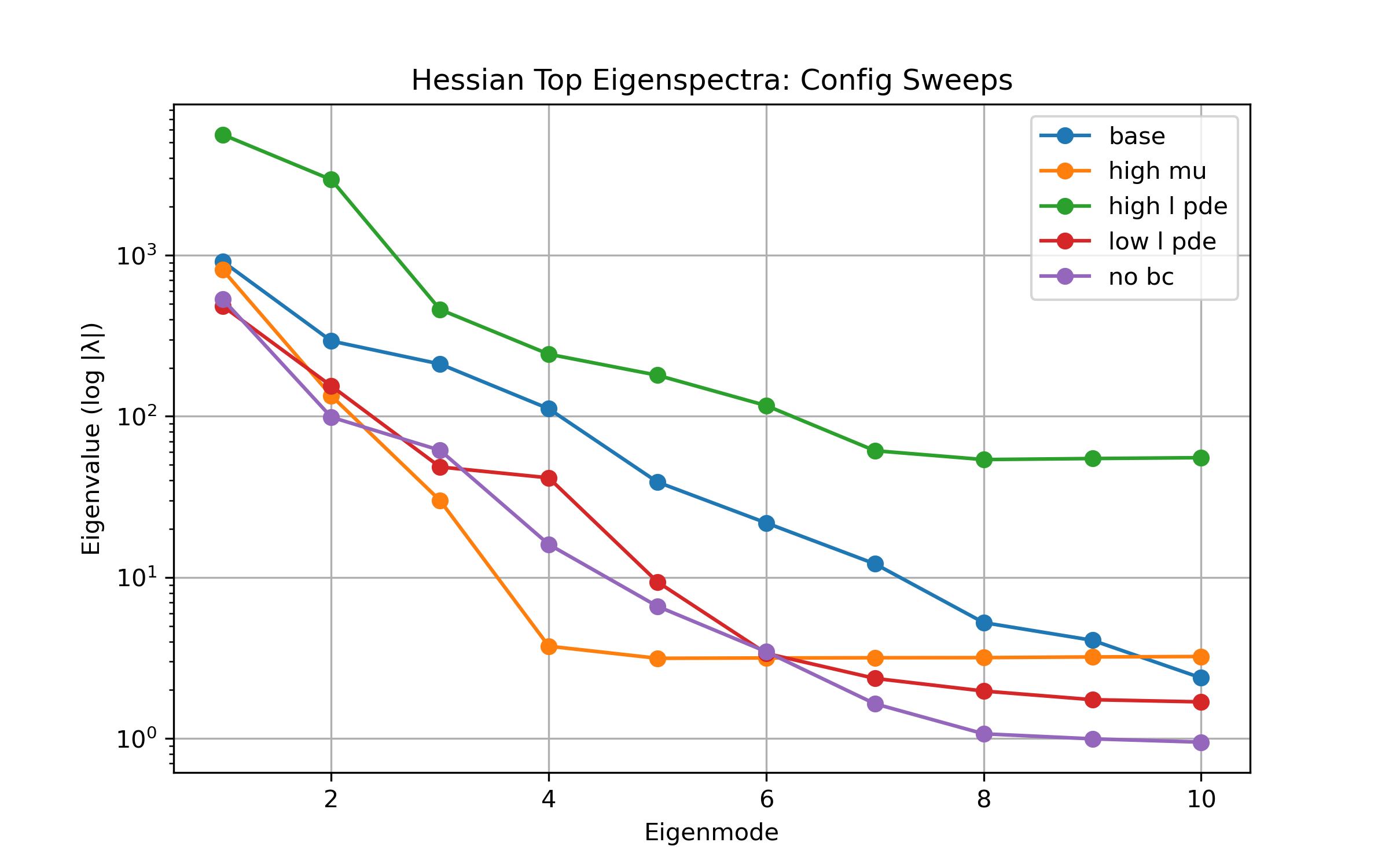}
\caption{Top eigenspectra of the total Hessian across configurations (log scale). High-$\mu$ produces the sharpest curvature, reflecting stiff dynamics.}
\label{fig:eigs}
\end{figure}

\begin{figure}[t]
\centering
\begin{subfigure}{.48\linewidth}
\includegraphics[width=\linewidth]{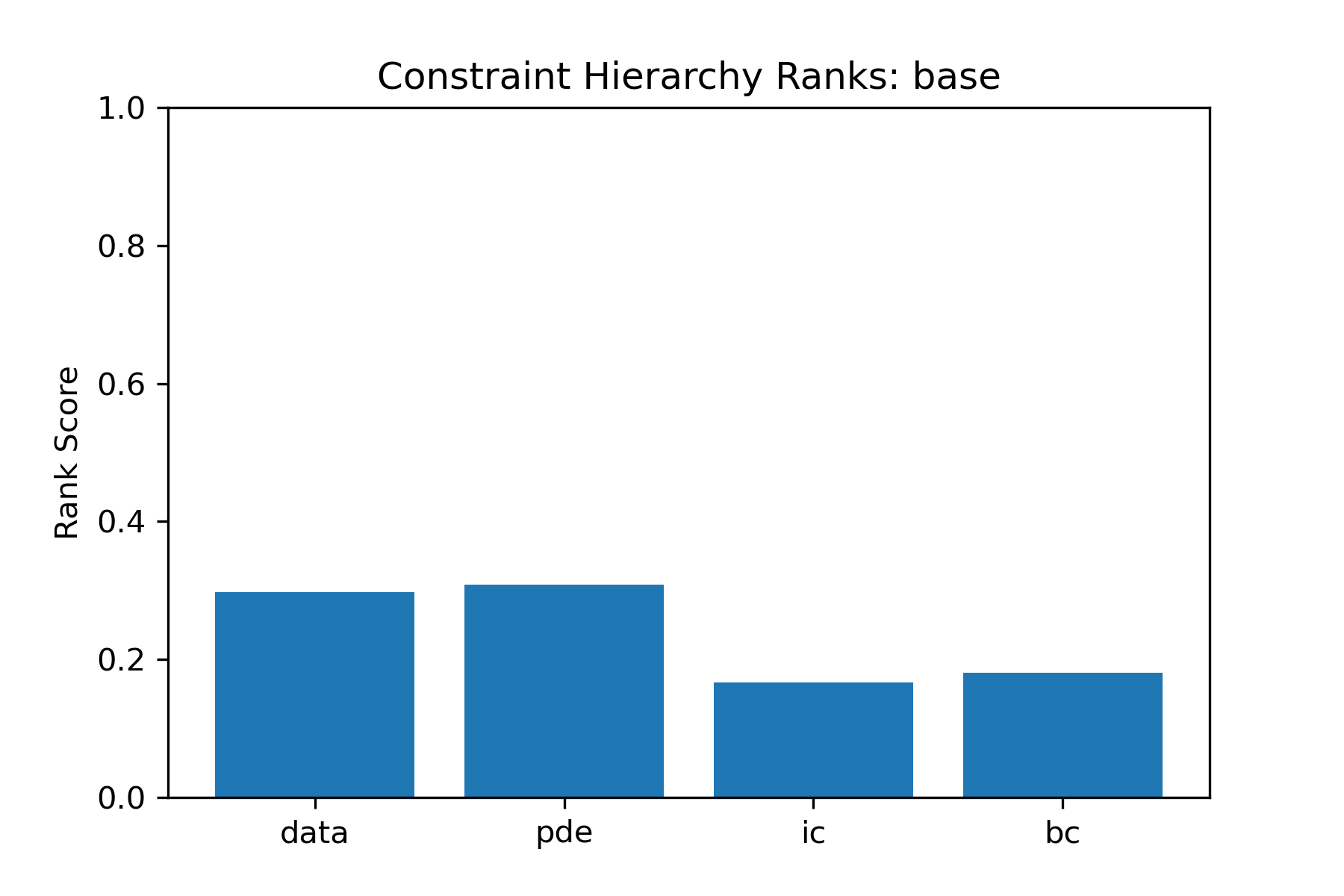}\caption{Base}
\end{subfigure}\hfill
\begin{subfigure}{.48\linewidth}
\includegraphics[width=\linewidth]{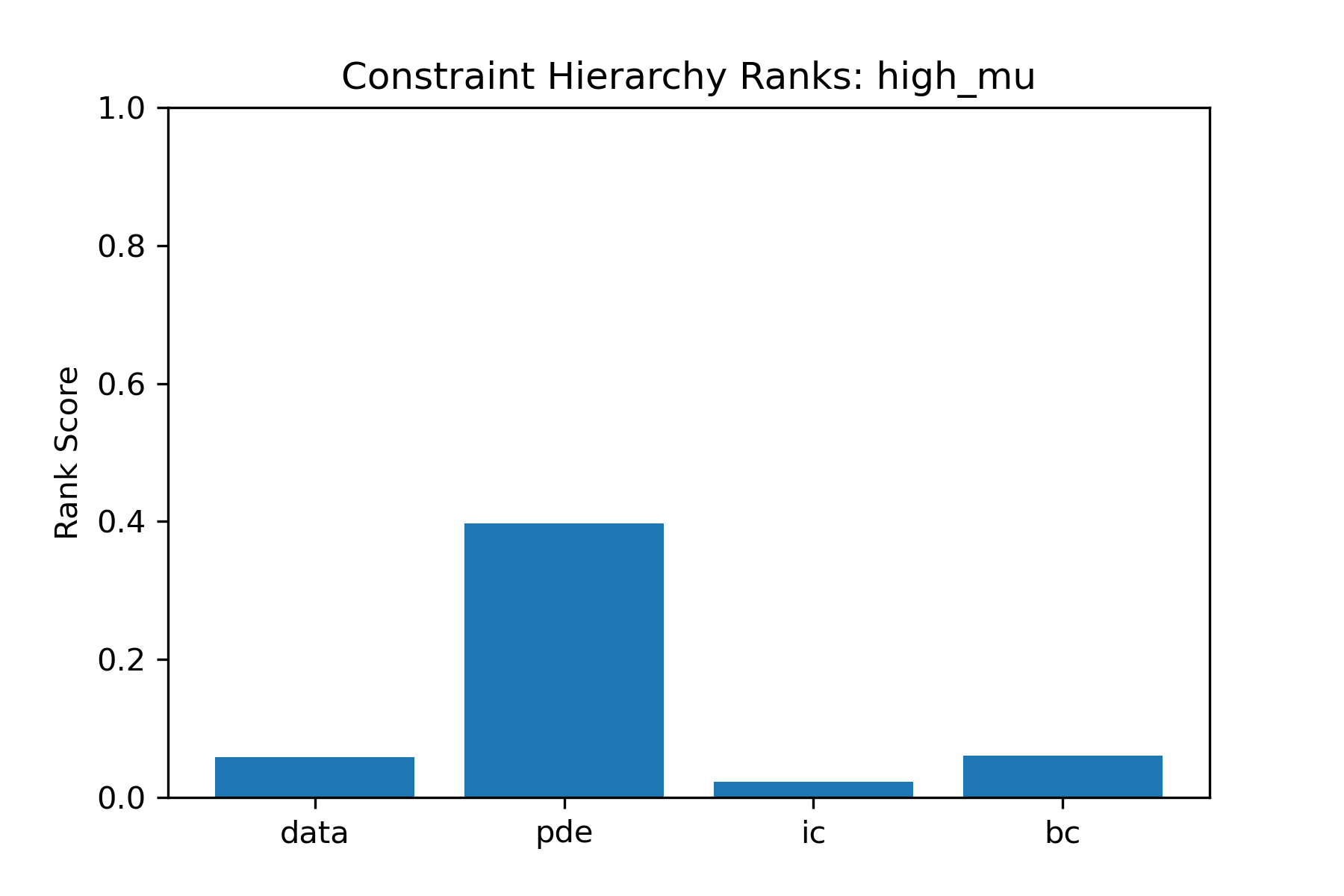}\caption{High-$\mu$}
\end{subfigure}

\begin{subfigure}{.48\linewidth}
\includegraphics[width=\linewidth]{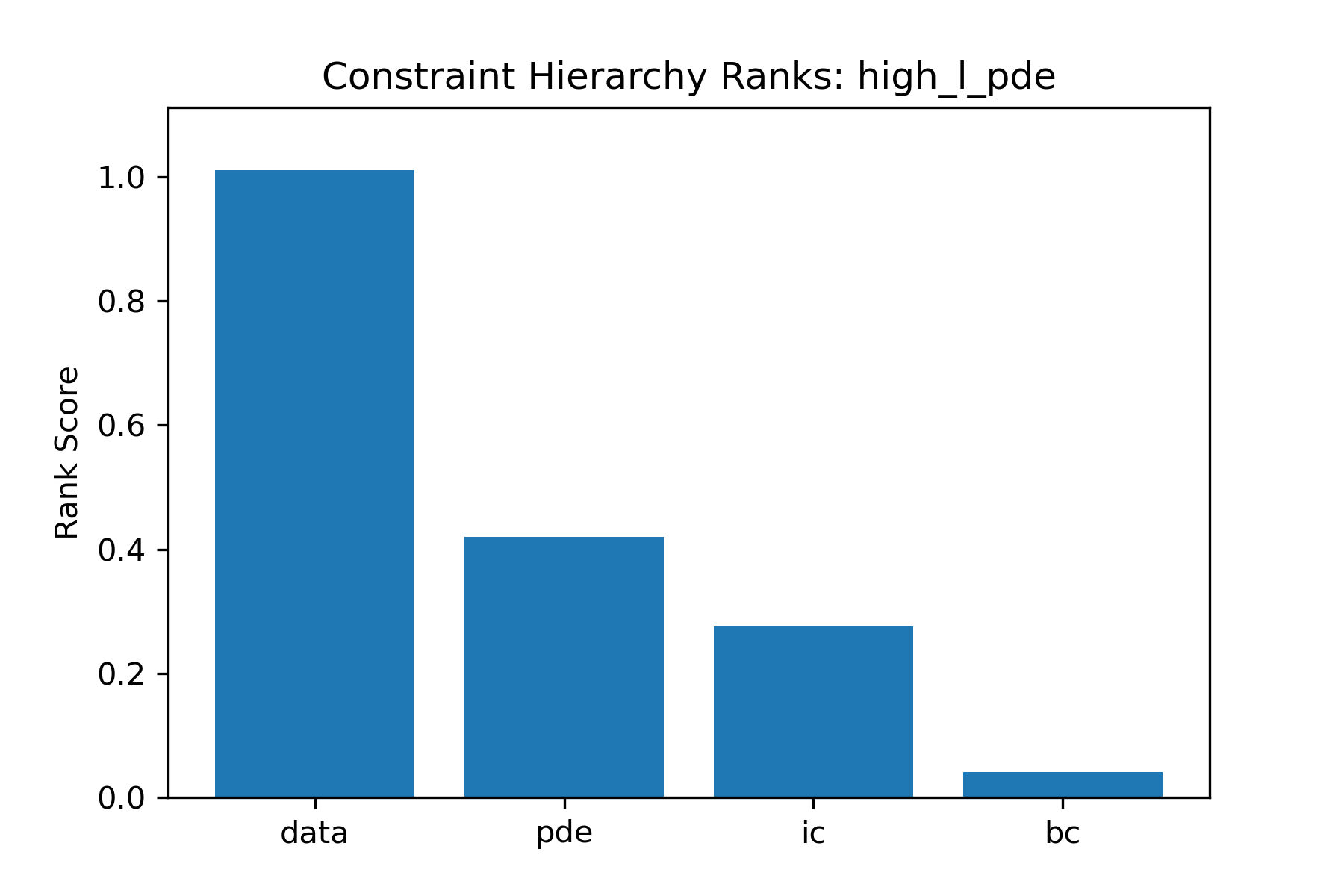}\caption{High-$\lambda_{\text{PDE}}$}
\end{subfigure}\hfill
\begin{subfigure}{.48\linewidth}
\includegraphics[width=\linewidth]{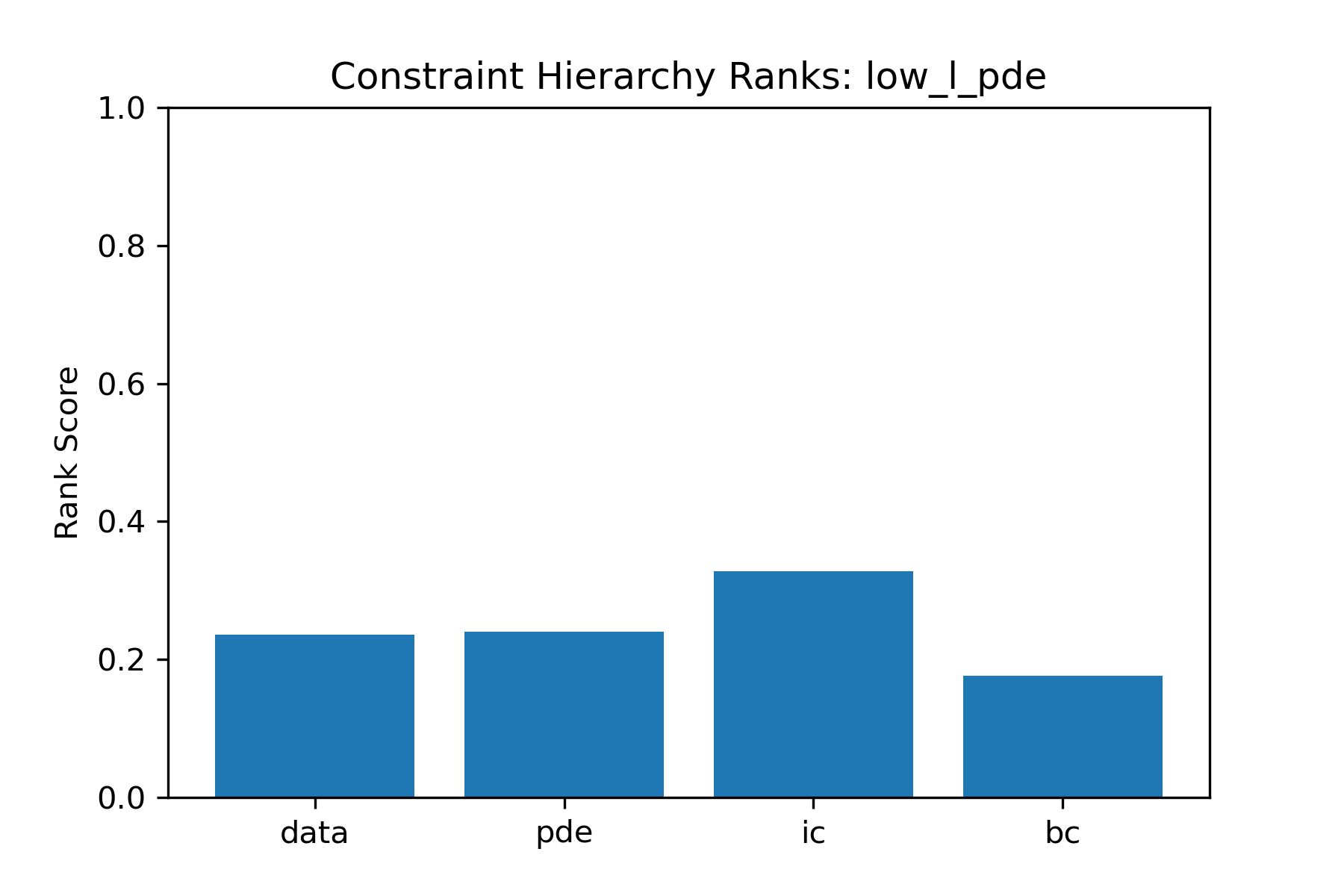}\caption{Low-$\lambda_{\text{PDE}}$}
\end{subfigure}

\begin{subfigure}{.48\linewidth}
\includegraphics[width=\linewidth]{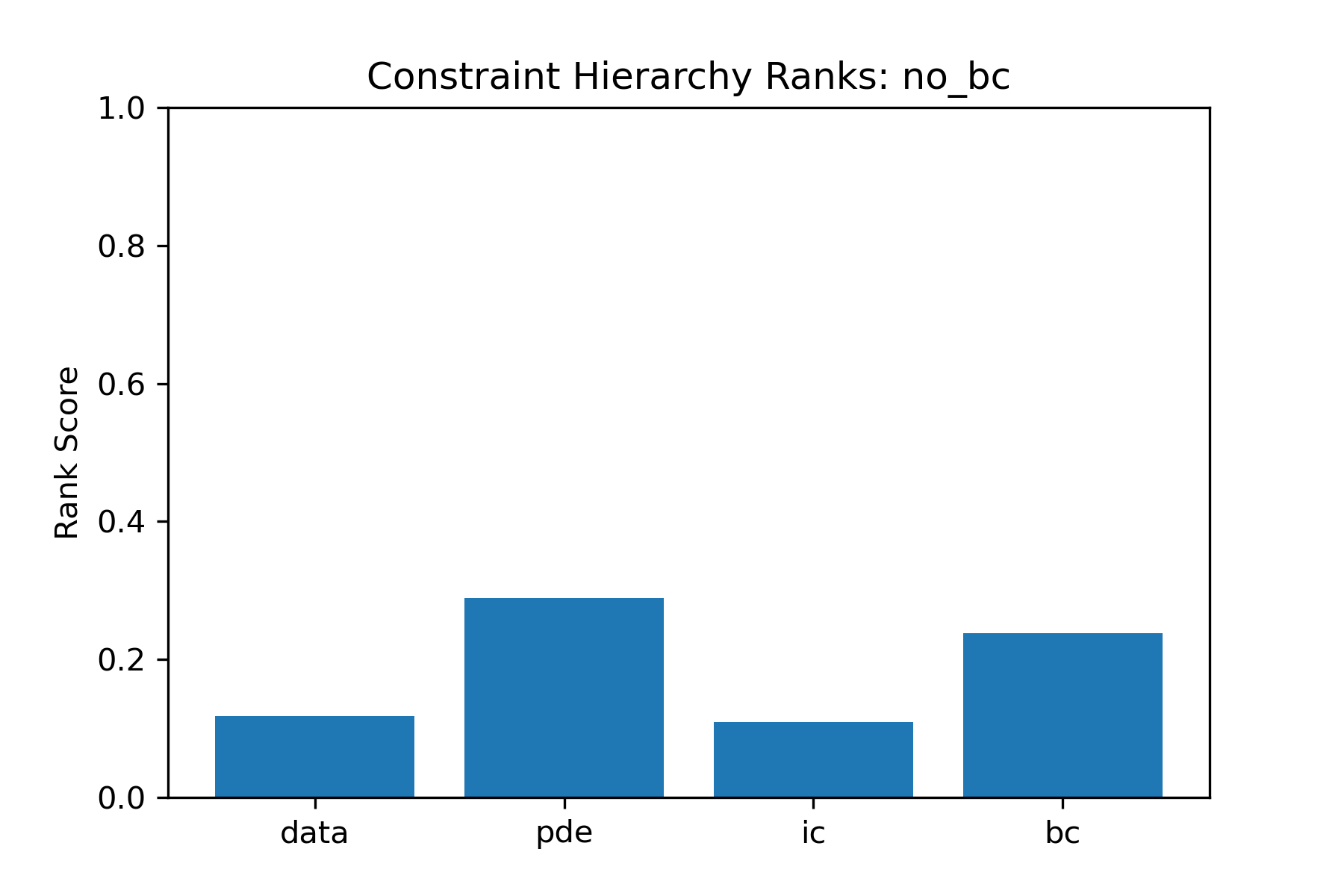}\caption{No-BC}
\end{subfigure}
\caption{Constraint hierarchy ranks across configurations. Higher bars indicate stronger curvature contribution after combining SC, AS, VA, and CNR.}
\label{fig:hierarchy}
\end{figure}

We applied the proposed framework to five trained B-PINN configurations and computed per-constraint metrics (SC, AS, VA, CNR), aggregating them into rank scores for visualization. Figures \ref{fig:eigs}  and \ref{fig:hierarchy} reveal coherent, physics-driven shifts in the constraint hierarchy that match physical intuition while exposing effects that naive loss‐weighting cannot capture.

In the base model, ranks are balanced: PDE and data contribute comparably ($\approx0.3$ each), with IC and BC slightly lower, providing a reference geometry.
Increasing the stiffness parameter ($\mu$) sharpens the dynamics: the PDE constraint dominates (rank $\approx0.4$) and drives the steepest Hessian eigenvalues (Figure \ref{fig:eigs}), consistent with the stronger conditioning of stiff oscillators.
Reducing the PDE weight ($\lambda_{\text{PDE}}$) redistributes influence toward the IC (rank $\approx0.4$), demonstrating that boundary-like information guides the posterior when physics is under-weighted.
Even when $\lambda_{\text{BC}}=0$, the BC term retains non-trivial rank ($\approx0.25$), indicating that the PDE implicitly enforces boundary consistency, a coupling that is invisible to simple weight inspection.

The high-$\lambda_{\text{PDE}}$ explicitly shows how loss term weighting do not guarantee dominance in curvature space. Despite the heavy PDE weighting, the data constraint achieves the highest rank ($\approx1.0$), and the PDE remains moderate ($\approx0.4$). Our spectral and alignment scores show that data gradients align more strongly with high-curvature directions, outcompeting the PDE in principal Hessian modes. 

We see explicitly in Figure \ref{fig:hierarchy} that constraints are non-linearly coupled through the network parameters, so varying one loss weight alters the effective influence of the others in a non-trivial way. Because the PDE, IC, and BC residuals all depend on the same set of parameters, their gradients span overlapping subspaces of the parameter space. Reducing, for example, $\lambda_{\text{PDE}}$, does not merely weaken the PDE term in isolation, it changes the geometry of the optimization landscape so that directions formerly constrained by the PDE become available to the IC or data terms, effectively amplifying their curvature contribution. Conversely, strengthening the PDE loss can indirectly suppress the gradient norms of the IC or data residuals by collapsing the solution manifold, even if their explicit weights are unchanged. Our hierarchy metrics capture this reallocation of influence: when one constraint’s weight is adjusted, the spectral contribution and alignment scores of the others shift in a manner that cannot be predicted by linear scaling alone. Typically, practitioners simply sweep over the weight space and pick the combination that yields the lowest validation error, remaining agnostic about how the constraints interact, but here we explicitly reveal how those interactions unfold. This highlights that the B-PINN objective is more than a weighted sum, the individual loss terms interact through shared Jacobians and higher-order curvature, so the realized dominance of each term emerges from the coupled geometry of the full system rather than from the loss weights themselves.

Across settings, steep leading eigenvalues coincide with suppressed predictive variance (low VA), further clarifying the origin of apparent overconfidence. For example, the high-$\mu$ case exhibits top eigenvalues exceeding $10^{3}$, and the Laplace variance $J_x^{\top}H^{-1}J_x$ contracts accordingly. Our per-constraint decomposition shows that this low variance is not a failure of Bayesian calibration but the natural consequence of physics-induced precision.

Prior PINN studies have visualized ill-conditioning or global spectra \cite{Krishnapriyan2021FailurePINNs, geniesse2024visualizinglossfunctionstopological,  Rathore2024PINNLandscapes}, but have not separated curvature by physical constraint, nor linked these measurements to Bayesian uncertainty. Our results demonstrate that the B-PINN posterior is shaped by an emergent hierarchy that can diverge sharply from the chosen loss weights, and that boundary effects can persist even when their loss terms are nominally absent. These findings give practitioners more equipment for identifying when low variance is physically warranted rather than an artifact of miscalibration.

\section{Conclusions and outlook}
In this work, we have introduced a methodological framework for quantifying constraint hierarchies in Bayesian physics-informed neural networks through per-constraint Hessian decomposition and empirically motivated metrics.
This perspective may quantify known imbalances in PINN training, where adjusting weights does not invariably ensure proportional dominance due to factors like gradient scales and problem conditioning \cite{wang2020understandingmitigatinggradientpathologies}. As illustrated in the Van der Pol regimes, we can see explicitly how over-weighting the PDE term may still yield data-dominated curvature, a phenomenon implicitly observed in prior parameter variations and adaptive schemes. In the near future it would be interesting to extrapolate more results from the Hessian, for a more families of PDEs. 

Standard caveats of the Laplace approximation apply: its local nature captures curvature within a single basin, potentially overlooking global effects that necessitates sampling-based validation. Furthermore, stiff regimes may require further numerical safeguards.

Understanding the effects of physical constraints on the network motivates curvature-informed adaptive weighting to enhance UQ robustness. On a broader level, advancing our grasp of the interplay between loss weights and physics-driven influences may demystify the black-box nature of neural networks, paving the way for future methods that extrapolate novel physical laws from trained models.

\begin{acknowledgments}
The work of FL is supported by STFC (ST/W507799/1). 
\end{acknowledgments}

\bibliographystyle{JHEP}
\bibliography{references.bib}


\newpage
\appendix


\end{document}